# Hate Content Detection via Novel Pre-Processing Sequencing and Ensemble Methods

**Anusha Chhabra[1], Dinesh Kumar Vishwakarma[2*]**
*anusha.chhabra@gmail.com[1], dinesh@dtu.ac.in[2*]*
*Biometric Research Laboratory, Department of Information Technology, Delhi Technological University, Delhi 110042, India*

**Abstract**
Social media, particularly Twitter, has seen a significant increase in incidents like trolling and hate speech. Thus, identifying hate speech is the need of the hour. This paper introduces a computational framework to curb the hate content on the web. Specifically, this study presents an exhaustive study of pre-processing approaches by studying the impact of changing the sequence of text pre-processing operations for the identification of hate content. The best-performing pre-processing sequence, when implemented with popular classification approaches like Support Vector Machine, Random Forest, Decision Tree, Logistic Regression and K-Neighbor provides a considerable boost in performance. Additionally, the best pre-processing sequence is used in conjunction with different ensemble methods, such as bagging, boosting and stacking to improve the performance further. Three publicly available benchmark datasets (WZ-LS, DT, and FOUNTA), were used to evaluate the proposed approach for hate speech identification. The proposed approach achieves a maximum accuracy of 95.14% highlighting the effectiveness of the unique pre-processing approach along with an ensemble classifier.

## 1. Introduction

Since Twitter is a well-liked social media site for sending short messages, it supports hundreds of millions of users in many different languages with restricted message length, making it difficult to identify hate speech [1]. Besides being a popular platform, Twitter is highly vulnerable to hate content. It is currently one of the most popular social media platforms for the automatic detection of in-text hate speech ([2], [3]) as well as a source of data for studies on abusive language. Due to the popularity of English and the fact that it is the most extensively used language and the most readily accessible data source [4], this research focused on messages posted on Twitter. Although popular social media platforms often delete the hate content before publishing them. Considering the fact, Twitter allows its users to decide what they see. The spreading of the content on Twitter depends on the retweeting, replying and liking of the posts by multiple users. As there are multiple definitions and variations of the hate forms available, the detection of hate content considering all forms is very difficult [5]. Due to the increase in the number of users globally, another challenge of detecting hate content on social media is the usage of diverse language posts and noisy data in terms of new terms adopted for communication [6], manual filtering of the messages is not possible. Therefore, there is a demand for automatic hate speech detection. It has also been observed that people also take the advantage of global pandemic situations for spreading hate content. During the initial phase of Covid-19 pandemic in India, #hatespeech was trending on Twitter platform [7]. There are uncountable hate posts available on social media platforms. Some of the examples on Twitter are as follows: "Let's kill Jews and kill them for fun." [8] and ''Twitter user Pu\*\*y a\*\* ni\*\*a'', and ''You hate football you are a fa\*\*ot.'' [9].
The above facts have motivated the researchers to focus on Natural Language Processing (NLP). NLP is one of the most demanding and promising field in computer research. Considering the NLP in text classification pipelines, preprocessing is proven to be one of the essential steps in various fields like Twitter sentiment analysis [10]. Using suitable text pre-processing methods or sequences makes it possible to raise the version of the classification problem. Preprocessing is to get text input ready for accurate computational method interpretation [11] and obtaining the high-performance classification models [12]. Previous studies investigating at how preprocessing affects text classification find varying outcomes for various preprocessing techniques and in various datasets [13].

The informal language on the web creates interference and noise that can mortify the performance of classification problems. According to the [14] survey, increasing noise by up to 40% in the tweet's dataset can affect the categorization problem's performance. This paper mainly focuses on the effect of text pre-processing, showing the importance of their sequence in identifying the hate content via ensemble techniques. Following this discussion, the effects of known feature extraction methods such as TF-IDF and count vectorization are also studied. The evaluation is done on the combination of three publicly available datasets ([15], [16], [17]). Contributions to this work have been mentioned below as follows:

•   Proposed a novel framework for identifying hate speech detection via considering the effect of data pre-processing.
•   This article presents an examination of various pre-processing methods, and the suitable selection of preprocessing sequence as the consequence of pre-processing step alone, the order in which they are applied, has not been studied on hate speech detection till now.
•   The suggested framework creates feature vectors by using a text augmentation and incorporating the count vectorizer library, which has been proven to be a good choice for feature extraction. Lastly, ensemble algorithms are trained on the combined datasets to detect hate speech in tweets using the built feature vectors and labels.
•   This research implements the ensemble approaches and shows the better accuracy on the selected preprocessing sequence.

The remaining portions of this study are outlined below: Section 2 discusses related works. In Section 3, the detailed methodology is presented covering the preprocessing methods, vectorization methods and ensemble techniques. In Section 4, the experimentation and result analysis are covered along with the dataset descriptions and hardware specifications. Section 5 covers the discussion followed by the conclusion and further future directions given in section 6.

## 2.    Related Work

We divide the relevant work into three subsections and present them as follows: The related work on text preprocessing techniques and on hate speech identification is introduced in section 2.1 and section 2.2, respectively. The related research on ensemble approaches is covered in sub section 2.3.

## 2.1.   Hate Speech Detection

Analyzing and extracting the emotions in text is a natural language processing application. The wide applications cover the sentiment analysis, hate speech detection, emotion detection, sarcasm detection. Hate speech detection is the most concerned research field as the number of users and their views are increasing day by day. The main challenge in the detection of hate speech is observed in [5], where authors clearly mentioned that there is no proper definition of hate speech, which makes the detection very cumbersome. In this paper, the work is mainly focused on English language and the research related to hate speech detection on the English language specifically is well resourced in ([18], [16], [19]). The primary method used to detect hate speech is binary text classification [20]. The research now a day also covers the detection of various hate forms like radicalization ([21], [22]), terrorism-based ([23], [24]), cyberbullying ([25], [26]), religion-based discrimination [27], racism and sexism [28].

## 2.2. Text pre-processing techniques

Text preprocessing plays a crucial role in text classification tasks. The effectiveness of various preprocessing techniques in automated text classification tasks have been the focus of many researchers in the past. The effect of preprocessing techniques can be seen in many real time applications like twitter sentiment classification ([29], [30], [31]), movie review [32]. A survey by [33] to improve short text quality also discusses the impact of preprocessing techniques before applying the models. Below, the work related to text pre-processing methods are discussed and the respective examples of tweets from hate speech dataset are also given:

**1.     Replacement of emojis and emoticons:** The users can show their moods and expressions through some text icons called emoticons. These text-based emoticons can be represented through some graphics commonly known as emojis. Social media users use numerous emojis and emoticons to express views and sentiments. If we classify our hate speech dataset with the opinion, these emojis contain helpful information. In an examination done by [34], these emojis and icons were substituted with their related word, such as "☹" this emoji replaced with the word "sad" and "☺" this replaced with the word "happy".

**Before:**
*Sunday is looking pretty good so far ☺*
**After:**
*Sunday is looking pretty good so far happy*

**2.     Eliminating URLs, noise, and hashtags:** The elimination adds noise to the data. Also, most tweets placed by users include URLs that mention extra valuable content for the user. The hashtag symbol also gives some additional information, sometimes expressing sentiment. This additional information is valid only for the user. This text doesn't provide any information to the machine. So, this information can be called as noise that needs to be managed. Many studies have already happened. According to a survey done by [35], these noisy things need to be substituted with other words, whereas a further study by [36] said we need to eliminate these noisy things.

**Before:**
*pussy is a powerful drug 😅 #HappyHumpDay http://t.co/R8jsymiB5b*
**After:**
*pussy is a powerful drug &128517 Happy Hump Day*

**3.     Word partition**: It is the activity of detaching the word/content utilized in a hashtag such as #usedbytrendypeople is divided as four words used + by + trendy + people. This activity is beneficial for machines to separate sentences into words without any human involvement.

**Before**:
*#usedbytrendypeople*
**After:**
*used by trendy people*

**4.     Substitute slang and abbreviation:** Every social media site limits the number of words in a comment. This restriction encourages online users to use short phrases and slangs. Abbreviations contain letters taken from the complete word, such as VIP stands for a significant person or BFN for a bye for now, BTW for, by the way, etc. Slang is also utilized as a casual mode of representing an idea and sometimes its point to a particular group of people. Hence it is crucial to deal with such informal words in the comments by exchanging them for their accurate word meaning. By doing this, the classifier's performance gets improved without information loss. In a survey done by [37], the text is well interpreted with standard text analytical tool if slang and abbreviation are substituted with their meaning.
**Before:**
*An article on: Money getting taller and bitches getting blurry*
**After:**
*Money getting taller and bitches getting blurry*

5. **Exchange extended Characters**: Social media users deliberately use extended speech where they intentionally compose or write many characters recurrent to give more importance, e.g., guuuuuuud, sweeeeeeeet. Therefore, if we do not convert these words into original words, the automated classifier thinks that these are different-meaning words, i.e., out-of-vocabulary (OOV). In our paper, we changed extended terms with their primary essential words. [38] performed experiments, discovered and changed these prolonged words, and their results showed that this replacement aided in tweaking the classification report.
**Before:**
*I just want some damn alone time. Fucccckkkkkk! Lol”*
**After:**
*I just want some damn alone time. Fuck! Lol”*

6. **Inaccurate spelling and grammar error**: They are usually existing in social media comments. Textblob2 and Norvig's spell correction methods is mostly used to correct the spelling and grammar.
**Before:**
*Experiencing first lounge for frst tym. Gud nyt.*
**After:**
*Experiencing first lounge for first time. Gud night.*

7. **Enlarge contractions:** This language is more informally written but mostly utilized by users to trim the letter counting. These short-form words are enlarged into their primary words so that machines can operate over them more easily. Such as well he'd, I'm, and I've are decomposable expressions that we need to change into their base word as we will, he would, I am, and I have, respectively. In the experiments done by [39], if we do not expand our short-form words into their original word, then the token created by the tokenization step is 'we' and 'll' from their contraction word is we'll, etc. and 'we' and 'll' words are meaningless.
**Before:**
*we don't pay hoes... we don't save hoes*
**After:**
*we do not pay hoes... we do not save hoes*

8. **Removing punctuation:** It is a way of adding more importance to our written language. Many people are not sure where and when to apply punctuation marks. Online users mostly use punctuations to reveal their thoughts, feelings, and opinions, but these are not helpful for the machine classifier. That's why we need to remove the punctuation marks in our data pre-processing step or give some way to replace these marks with words that express the same meaning. In the study of [40], they delete punctuation marks. Still, in another study by [41], they provided another approach where they didn't remove them but replaced them with some appropriate tags like this '!' mark replaced with the word 'surprise.'
**Before:**
*!!!!!!!!!she look like a tranny*
**After:**
*She look like a tranny*

**9. Delete unwanted numerals:** Online users also use numbers we don't need, giving some ambiguity to automated machine classifiers. In the study presented by [42], the authors deleted all numbers from the text. But by doing this, sometimes meaningful data is also lost. So first, we need to replace slang and short-form word, and then we can remove unwanted numbers. For example, words like 2m2h, 2nite, and b2w are replaced with base words like too much to handle, tonight, and back to work.
**Before:**
*I just want some damn alone time. Lol82211*
**After:**
*I just want some damn alone time. Lol*

**10. Case-folding:** It's a way of replacing all alphabets with a small case. By doing this step, we remove ambiguity due to their upper and lower case. This ambiguity creates a problem for the machine classifier. Replacing all upper case to lower case is the most common process for dealing with cryptic text problems. But sometimes, it creates a situation like in the US; if we convert it to lowercase, it will become 'us' before it was the country name. In the study of [43], they explained all the problems while we converted all upper cases into lower cases.
**Before:**
   *Go to the airport early. How can I change the flights.*
**After:**
   *Go to the airport early. how can i change the flight.*

**11. Delete stop-words**: These are a group of usually used letters in a speech. Words like is, in, the, etc. We need to remove that stop words because they sustain minimal informative data and are not helpful in machine classifiers.
**Before**:
   He plays cricket, and cheat on girls.
**After:**
   He plays cricket, and cheat on girls.

**12. Lemmatization:** Written context may contain words like 'utilise' and' utilize', but these are derived from one base word like 'use.' These derivable words have a similar meaning. In the stemming technique, cut off the end of the term. But in lemmatization, we use vocabulary and knowledge to convert a word into its original word. Many libraries are available in python to attain this lemmatization, like nltk, CoreNLP 8, and TextBlob 9.
**Before:**
*running round here like some brand new pussy thats bout to get fucked*
**After:**
*Run round here like some brand new pussy that bout to get fuck*

### 2.3.    Ensemble Approaches in Machine Learning
This section discusses the related work done in ensemble methods. As of now, there are three ensemble approaches: Bagging, Boosting and Stacking. While the base classifiers are trained on various subsets of the training patterns, the greatest set of ensembles creates ensemble classifiers by altering the training patterns.

The subsets are generated using various approaches. The training subsets are generated at random (with replacement) from the training set in bagging [44]. The subsets are used to train homogeneous base classifiers. The class selected by the majority of base classifiers is regarded as the ensemble classifier's final determination. Random forests [45] and large-scale bagging [46] are two examples of the many variations of bagging and aggregation procedures. By resampling the training patterns, boosting [47] builds data subsets for base classifier training while only using the most instructive training pattern for each succeeding classifier. Boosting is more broadly used in AdaBoost [48]. Stacking [49] with meta level features is proven to the best classifier.

## 3.  Methodology

This section describes the analysis of various pre-processing methods and the framework for the proposed Ensemble models to find out hatred speech, as represented in Fig 1. From the three-tier proposed architecture, level 0 specifically intended to data processing and feature extraction phase. The base module, Level 1, is where the data is split up into training categories, validation and testing folds (80:10:10) before passing to the base models. Level 1 is also utilized to generate prediction for creating new data. that Here, we examine the pre-processing methods used to elaborate their effect on tweet text. Lastly, Level 2 is the combined module in which three ensemble techniques like bagging, boosting and stacking are performed to determine the labels as hate or non-hate.

### 3.1.  Suggested Sequence of Pre-Processing Methods:

The primary purpose for doing this process is to refine the quality of the text and discover a collection and series of pre-processing methods that execute best in comparison with others. Many employs four to five ordinary pre-processing methods; after that, they use any classification model. But we need to apply more pre-processing techniques when dealing with twitter comments to normalize and improve text quality. Eliminating URLs, and hashtag symbols and replacing short-form words, emojis, and contractions with their original word are practical steps for automated text analysis. The choice of proper order and combination of the pre-processing method is the primary step to improve the classifier's performance. Some techniques perform well when used individually but not with other styles, and some aggregation of pre-processing methods does not act effectively. If we choose the wrong combination of techniques, it may degrade the classifier's performance due to information loss. Wide varieties are possible from these 12 preprocessing techniques. So, we used the following pre-processing sequence to tweak the classifier's performance.

In order to attain an improved quality converted tweet, some techniques are #I (Elimination of URL, noise, hashtags), #IX (Delete unwanted numerals), #XI (Delete stop-words), and #XII (Stemming/Lemmatization) have to be performed in the same sequence, and thus bring down the achievable number of aggregations of the leftover pre-processing methods to T8. As explained, it is the choice of a combination of essential pre-processing methods and their sequence. For example, if we remove the number 2 from the word '2nite' before exchanging it with the actual word 'tonight,' the outcome is message loss. Similarly, if we do first tokenization and then apply enlarge contraction also impacts performance. If we first do tokenization for a word like "w'll" then this word is converted into w and ll, and then we apply to enlarge contraction, it is not possible to expand this word as "we" and "will." So, the sequence of pre-processing should be like this; first, we have to do 'remove hash tag' followed by significant contraction, and finally, word segmentation or tokenization, then "# w'll" is correctly expanded to "we" "will." With the above stated need in mind, we research with different sequences. Only those sequences that showed remarkable performance are shown in Table 1.

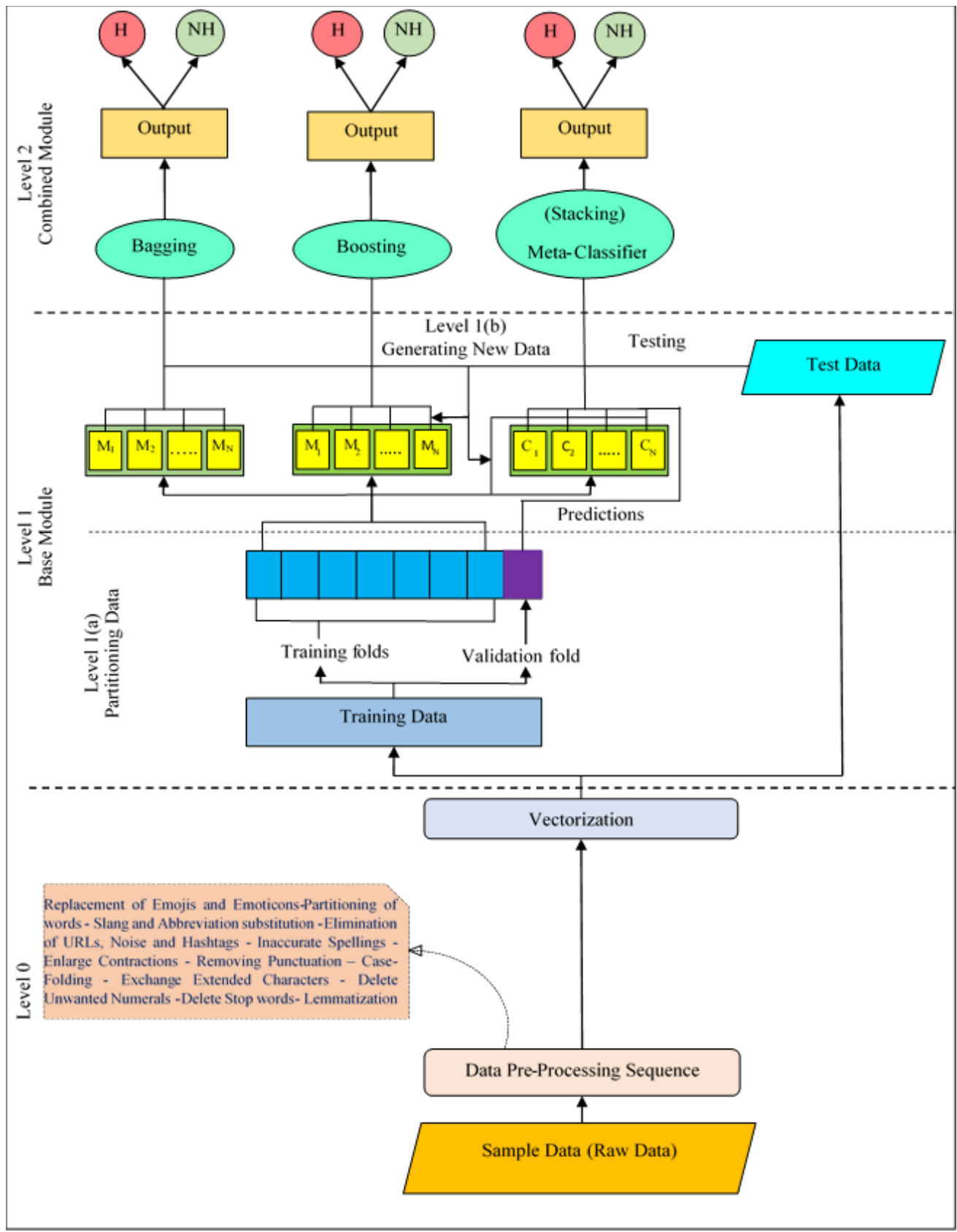

Fig 1. Proposed Architecture

Table 1 Steps of Pre-processing Sequence

| S No | Sets for pre-processing sequence | Roman- Numerals connected with preprocessing methods | Alias (for reference) |
|---|---|---|---|
| 1 | 1-2-3-4-5-6-7-8-9-10-11-12 | I-IV-VI-IX-III-XII-VIII-V-X-XI-VII-II | T1 |
| 2 | 1-12-5-2-8-3-11-7-4-9-10-6 | I-II-III-IV-V-VI-VII-VIII-IX-X-XI-XII | T2 |
| 3 | 1-8-5-2-11-12-3-7-4-9-10-6 | I-V-III-IV-VII-II-VI-VIII-IX-X-XI-XII | T3 |
| 4 | 1-2-11-8-12-5-3-7-4-9-10-6 | I-IV-VII-V-II-III-VI-VIII-IX-X-XI-XII | T4 |
| 5 | 1-8-11-12-5-2-3-7-4-9-10-6 | I-V-VII-II-III-IV-VI-VIII-IX-X-XI-XII | T5 |
| 6 | 1-8-2-5-11-12-3-7-9-4-10-6 | I-V-IV-III-VII-II-VI-VIII-X-IX-XI-XII | T6 |
| 7 | 1-2-5-12-3-11-7-9-8-4-10-6 | I-IV-III-II-VI-VII-VIII-X-V-IX-XI-XII | T7 |
| **8 (Proposed)** | **1-5-2-12-3-11-7-9-8-4-10-6** | **I-III-IV-II-VI-VII-VIII-X-V-IX-XI-XII** | **T8** |

In Table 1, we have shown the series of steps that we followed for pre-processing; first, we converted our input data into clean data, removed unwanted data, and then applied tokenization to convert the whole message into words. After that, we remove stop words that do not add value to our data. We used a porter stemmer for stemming. Stemming is the process of reducing derived words into their root word, as coding changed to code, but at the same time, we have to keep one thing in mind don't over stemming and under stemming; otherwise, its meaning gets lost. By stemming, we can relate the word with similar meanings. The next step is lemmatization performed better to find out the root word that has some dictionary meaning.

### 3.2. Words representation

In this step, we will examine which feature extraction method is better and can be used to tweak the classifier's performance. Word representation is the method where input is pre-processed data, and output is the numerical representation of this data that can be used in the classifier because the classifier can work only on numeric data.

- **Count Vectorization**: It is a process for transforming a given cleaned message into the frequency of each word. It checks how many times this word occurred in a clear message. As shown in Fig 2, Text1 does not include the words "dog" and "monkey"; hence, their oftenness is measured as zero while other words have frequencies of one. However, it has some problems: it does not explain which word is crucial for analysis.

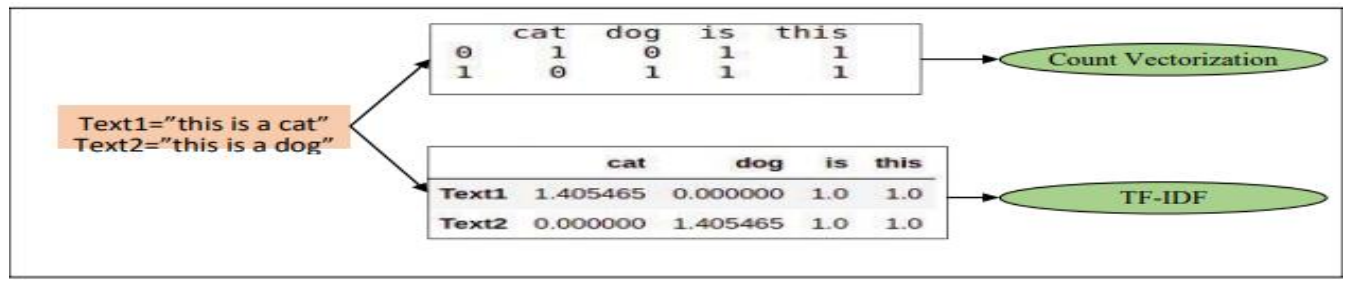

Fig 2. Vectorization Process

- **TF-IDF**: It also calculates the term frequency on the word counts. It also gives information about which word is more important than the other does. Therefore, by using this information, we can delete those less critical words, which affects our classifier if we reduce the dimension of input data.

The pre-processed data is transformed into numbers using both vectorizers.

## 3.3. Ensemble Techniques

In this phase, the output of the previous step is a numeric representation of text used to train classifiers like Support vector machine (SVM), Decision Tree (DT), logistic regression (LR), K- Neighbor classifier (KN) Random Forest (RF) to detect hatred speech. Afterward, the same data is also used to train the classifier based on ensemble learning techniques like the bagging, boosting, and stacking classifier. In ensemble techniques, we have used several classifiers. Therefore, we initially check for the base model which can give the best performance on the given dataset, and then we can ensemble these best base models to make our new model. That proposed model is used to give the result.

In the bagging ensemble technique, we collect the outcomes of base or weak models to get the result. However, if all base or weak models take the same data, they will be useless. We have implemented bootstrapping to create subsets from the input dataset with some replacement, but the size of each subgroup is the same as the input dataset. After that, the weak model is defined for each subset so that these weak models can run together but independently of each other. The results are obtained by collecting the outcome from all weak models. Boosting is also a good ensemble technique that proves good results; here, each weak model tries to remove the fault of the previous weak model. In the first step data subset is created from the original dataset. A weak model is defined on these subsets same as bagging. Faults are calculated using the original value and prediction outcome. Here fault predictions give higher weights. The following base model tries to remove the previous fault. The final proposed or robust model is created by the mean of all the weak or base models. So, the boosting algorithm collects a variety of weak model and finally make a strong model because each weak model increases the performance of the final strong model. Stacking is a technique that takes help from the outcome of various models (SVM, logistic regression, KNN, etc.) to construct a new model. This proposed model usually makes predictions on the test dataset.

The following sub sections presents the algorithms of all the ensemble approaches.

### 3.3.1. Bagging:

In the Bagging algorithm, $'N'$ samples of tweets with matching binary labels make up the input training set $B$ for Bagging. Lines 2–5 draw a bootstrap sample $D_j$ with replacement $'N'$ times from the training set $B$ at the first iteration when $k = 1$. An algorithm for machine learning is then given $D_j$ to induce model $'h'$ with classifier $C_j$. Lines 2 through 5 are repeated an additional $k - 1$ times, producing $k$ model iterations. The classification of a new tweet using the $k$ model majority vote is done in classification pseudocode, where I (.) is an indicator function that produces 1 if the input arguments are true. If not, it produces nothing.

**Algorithm 1** Bagging Algorithm (**Classifier Generation**)

1: **Bagging** $(B = \{(x_1, y_1), (x_2, y_2) \dots \dots \dots (x_N, y_N)\})$
　　　Where, N be the size of the training set.
2:　$for\ j = 1\ to\ k\ do$
3:　　Draw a bootstrap sample:
　　　$D_j(size: N) \Rightarrow B$ ⎬ *Sample of instances to input forming the original sample*
4:　　Apply the learning algorithm to $D_j$
5:　$end\ for$
6: Storing the classifier $C_j$ with Inducer $'h'$

Bagging Algorithm (**Classification**)

1: $C^*(x) = argmax\ \#\ \{I(j: C_j(x)) = y\}$
2: $y \in Y(hate, non\ hate)$
　　Where, $I(.) = 1\ if\ the\ expression\ is\ true\ and\ 0\ otherwise$

### 3.3.2. Adaboost:

AdaBoost initially receives a training sample made up of $'N'$ tweet instances, as can be seen from Algorithm 2. Therefore, in the first iteration, all instances are allocated equal weights recorded in $\varphi 1$ (Lines 2–4). We carry out the following in lines 5–14 of the first iteration: To generate the training set D1, we took $'N'$ samples from A using the weights in and fed it to the machine learning algorithm (lines 6–7). The importance and error of hypothesis h1 are then determined (Lines 8–9). The weight of poorly categorized examples is updated in lines 10–12 to receive greater consideration in the following iteration, while the weight of correctly identified examples is decreased. These steps (i.e., lines 5–14) are repeated an additional k-1 times, yielding k. We use a weighted majority vote to make a prediction for a new tweet. (classification pseudocode).

**Algorithm 2** Adaboost Algorithm (**Classifier Generation**)

1: **Adaboost** $(A = \{(x_1, y_1), (x_2, y_2) \dots \dots \dots (x_N, y_N)\})$
   Where, N be the size of the training set.
2:  $for\ i = 1\ to\ N\ do$     // *Assigning equal weights to each of the data points*
3:      $\varphi_1 \leftarrow \frac{1}{N}$
4:  $end\ for$
5:  $for\ j = 1\ to\ k\ do$
6:      Create a training set for A base on $\varphi_j$
7:      Apply the learning algorithm to $D_j$
8:      Get weak hypothesis $h_t: X \rightarrow \{-1,1\}$ with error $\in_j = \Sigma i: h_j(x_i) \neq y_i \varphi_j(x_i)$
9:      Choose $\alpha_j = \frac{1}{2} \log \log (\frac{1-\in_j}{\in_j})$
10:     Update:
11:     $for\ i = 1\ to\ N\ do$
12:         $\varphi_{j+1}(i) \leftarrow \frac{\varphi_j(i) exp\ (-\alpha_j h_j(x_i) y_i)}{Z_j}$
        Where, $Z_j$ is a Normalization factor (Chosen so that $\sum_{i=0}^{N}(\varphi_{j+1} = 1)$
13:     $end\ for$
14: $end\ for$
15: Storing the classifier $C_j$ with Inducer $'h'$

Adaboost Algorithm (**Classification**)

1: $C^*(x) = sign\ \#\ \{I(j: C_j(x)) = y\}$
2: $y \in Y(hate, non\ hate)$
   Where, $I(.) = 1\ if\ the\ expression\ is\ true\ and\ 0\ otherwise$

### 3.3.3. Stacking:

Stacking Algorithm initially receives a training set made up of $'N'$ tweet instances, as can be seen from Algorithm 3. Therefore, in the first iteration, all instances are running as a base learner classifier C1 (Lines 2–4). To generate new datasets from S, the results of many base learners are generalized by meta-learners because the low-level output is used as the input of the high level for relearning. It is carefully monitored to make sure that the level-0 examples are not used to train the meta-learner in order to prevent the overfitting issue (lines 5–9).

For a new tweet, second level model is utilized to generate prediction (classification pseudocode).

**Algorithm 3** Stacking Algorithm (**Classifier Generation**)

1: **Stacking** $(S = \{(x_1, y_1), (x_2, y_2) \dots \dots \dots (x_N, y_N)\})$
   Where, N be the size of the training set.
2:  $for\ i = 1\ to\ N\ do$     // *Learn $1^{st}$ level classifier*
3:      Learn a base classifier $C_i$ based on S
4:  $end\ for$
5:  $for\ j = 1\ to\ k\ do$     // *Construct new datasets from S*
6       $\{\acute{x}_j, y_j\},$

|   | Where, $\acute{x}_j = \{C_1(x_j), C_2(x_j), \ldots, C_i(x_j)\}$ |
|---|---|
| 7: | end for |
| 8: | Learn a new classifier $C_j$ based on steps 5-7 |
| 9: | Storing the classifier $C_j$ with Inducer $'h'$ |

| Stacking Algorithm (**Classification**) |
|---|
| 1: $C^*(x) = \{I\{C_j(C_1(x), C_2(x), \ldots, C_i(x)) = y\}$ |
| 2: $y \in Y(hate, non\ hate)$ |
| Where, $I(.) = 1\ if\ the\ expression\ is\ true\ and\ 0\ otherwise$ |

## 4. Experimentation and Result Analysis

We conduct experimentation on the combination of three publicly available datasets. This section provides the brief dataset descriptions which are used for the combination, various classification metrics used for evaluation of results, hardware requirements.

### 4.1. Dataset Description

We test our model using three publically available datasets. Three datasets, WZ-LS [15], DT [16], and FOUNTA [17], are largely employed in studies on the identification of hate speech. The three datasets are joined, and the spam tweets are removed, to create the combined dataset. The summary of these datasets is shown in Table 2. The table clearly shows the wide range in size and quality of different datasets, allowing us to assess our suggested model more thoroughly.

#### 4.1.1. WZ-LS dataset

The combination of two Twitter datasets ([19], [50]) are used to make the WZ-LS dataset [15]. The dataset splits down the hate speech class into sexism, racism, and neither. They recover the text of the tweets using the Twitter APIs, but Twitter has removed a few due to their inappropriate content.

#### 4.1.2. DT dataset

Authors in [16] observed that hates speech should be distinguished from offensive tweets; several tweets may include hateful words classified as offensive and do not match the threshold to categorize them as hate speech. The investigation rebuilt the DT Twitter dataset, which was manually tagged and classified into three categories: offensive, hate, and neither.

#### 4.1.3. FOUNTA dataset

The FOUNTA dataset [17] is a human-annotated dataset; first-time annotators categorize tweets into three classes: normal, spam, and inappropriate. The last variant of the dataset has four categories: normal, spam, hate, and abusive.

Here, we eliminate the identical resulting in the distribution in Table 2.

Table 2 Abstracts of Datasets

| Ref | Dataset | Size | Labels |
|---|---|---|---|
| [15] | WZ-LS | 13,202 | racism (82), sexism (3,332), both (21), neither (9,767) |
| [16] | DT | 24,783 | hate (1,430), offensive (19,190), neither (4,163) |
| [17] | FOUNTA | 89,990 | normal (53,011), abusive (19,232), spam (13,840), hate (3,907) |
| - | Combined | 114,120 | normal (66,941), inappropriate (47,194) |

### 4.2. Classification Metrics

The results are evaluated in terms of accuracy, precision, recall, F1- Score. Table 3 shows the general formulas for the above-mentioned performance metrics along with their ranges.

Table 3 Classification Metrics

| Metrics | Formula | Range |
|---|---|---|
| Accuracy | $\dfrac{TP + TN}{TP + TN + FP + FN}$ | [0,1] |
| Precision | $\dfrac{TP}{TP + FP}$ | [0,1] |
| Recall | $\dfrac{TP}{TP + FN}$ | [0,1] |
| F1-Score | $\dfrac{(2 \times Recall \times Precision)}{Recall + Precision}$ | [0,1] |

### 4.3. Hardware

We have conducted the proposed experiment on NVIDIA Titan RTX 24GB GPUs and the system memory used is 128GB.

### 4.4. Results and Discussion

The experiments are done on the combined dataset initially by considering the various base models like logistic regression, decision tree, support vector machines and K-Neighbor classifiers on all the pre-processing sequences (refer Table 1) but the main focus of the paper is on the ensemble technique on the proposed sequence i.e., T8 (Table 1) because it combines various models' results to make a robust model, clearly showing the effect of pre-processing sequence.

Table 4 Accuracy Scores of base models on all aliases (refer Table 1)

| Models | Accuracy Scores | | | | | | | |
|---|---|---|---|---|---|---|---|---|
| | T1 | T2 | T3 | T4 | T5 | T6 | T7 | T8 (Proposed) |
| **SVM** | 0.9299 | 0.9423 | 0.9225 | 0.9358 | 0.9444 | 0.9335 | 0.9246 | **0.9480** |
| **KN** | 0.9222 | 0.9284 | 0.9119 | 0.9302 | 0.9322 | 0.9269 | 0.9158 | **0.9380** |
| **LR** | 0.9011 | 0.9006 | 0.8965 | 0.9156 | 0.9225 | 0.9147 | 0.9112 | **0.9289** |
| **DT** | 0.8845 | 0.9121 | 0.9089 | 0.9111 | 0.9301 | 0.9232 | 0.9202 | **0.9320** |
| **RF** | 0.9025 | 0.9288 | 0.9236 | 0.9312 | 0.9356 | 0.9300 | 0.9299 | **0.9368** |

As T8 (proposed) sequence is giving good accuracy scores (see Table 4). Therefore, other performance metrics like precision, recall, F1-measure are evaluated on T8 only (Table 5).

Table 5 Performance of Base Models on T8 Sequence

| S.No | Models | Precision | Recall | F1 -Measure | Accuracy |
|---|---|---|---|---|---|
| 1. | **SVM** | 0.9600 | 0.9800 | 0.9698 | 0.9480 |
| 2. | **K-Neighbour** | 0.9400 | 0.9900 | 0.9643 | 0.9380 |
| 3. | **Logistic Regression** | 0.9500 | 0.9900 | 0.9696 | 0.9289 |
| 4. | **Decision Tree** | 0.9400 | 0.9900 | 0.9643 | 0.9320 |
| 5. | **Random Forest** | 0.9400 | 0.9900 | 0.9643 | 0.9368 |

Fig 3 presents the graphical representation of the performance scores of base models on T8 in terms of precision, recall, F1and accuracy.

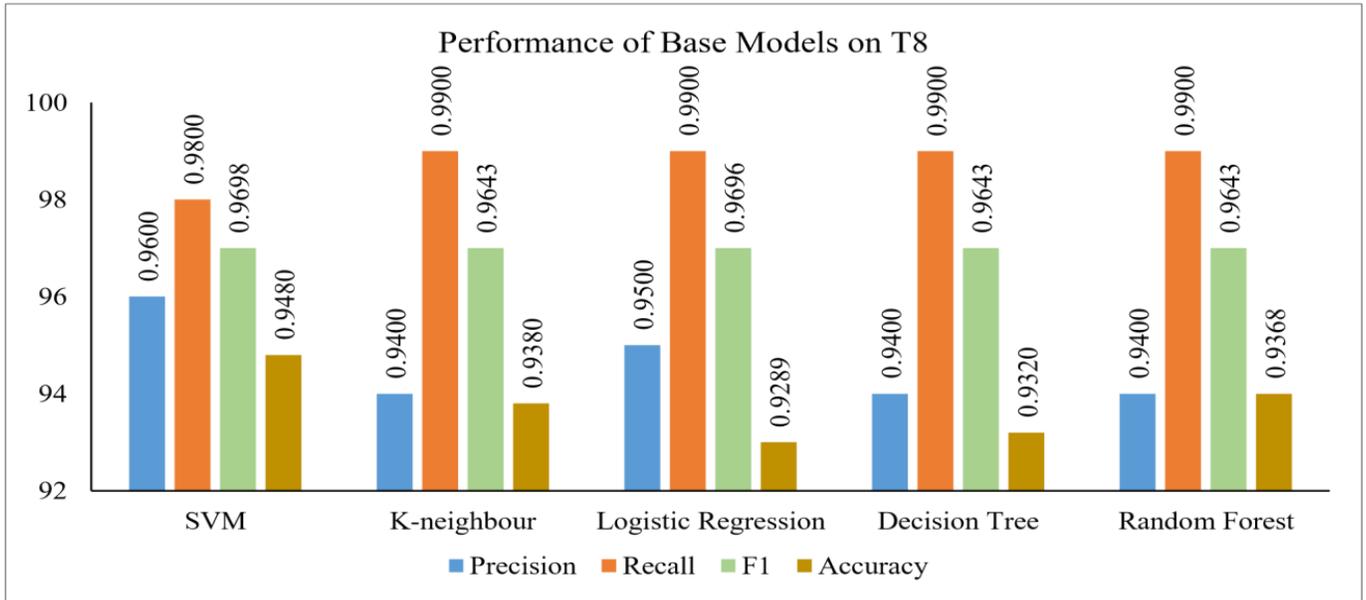

Fig 3 Graphical Representation of the Performance scores on T8

After the obtained scores from the base models, we applied an ensemble technique that use the best base model to tweak the performance. In the bagging classifier ensemble technique, by default, it takes random forest as the base model because it gives results from a number of the decision tree as a base model. After that, we tried different weak models like K-neighbour and logistic regression that increased the performance to 0.9490. The Adaboost classifier ensemble technique uses the base model as logistic regression because that was the best model when we tested individually; this proposed model increased performance up to 0.9350. And finally, we tried Stacking ensemble method, the meta classifier. Here, we tried two types of voting methods that are hard and soft, but both of them uses the same combination of three base models, i.e., logistic, SVM, and random forest. In hard voting, the outcome will be the mode of output labels with equal weights. If the meta classifier in the hard vote has unequal weight, then apply that unequal weights over the outcome of labels and give the final result. The result for the soft meta classifier will be based on the probabilities of all the outcomes from various weak models. Table 6 shows the performance of our proposed approach.

Table 6 Accuracy Scores of Ensemble Techniques

| **Models** | | **Accuracy** |
|---|---|---|
| **Bagging Classifier** | Random Forest K-Neighbour Logistic Regression | 0.9410 |
| | | 0.9404 |
| | | 0.9480 |
| **Boosting Classifier (Adaboost)** | Decision Tree (Default) Logistic Regression | 0.9350 |
| | | 0.9358 |
| **Stacking Classifier (Hard)** | Logistic Regression (Base Model), SVM, Random Forest | 0.9514 |
| **Stacking Classifier (Soft)** | | 0.9512 |

Fig 4 shows the pictorial representation of our proposed approach.

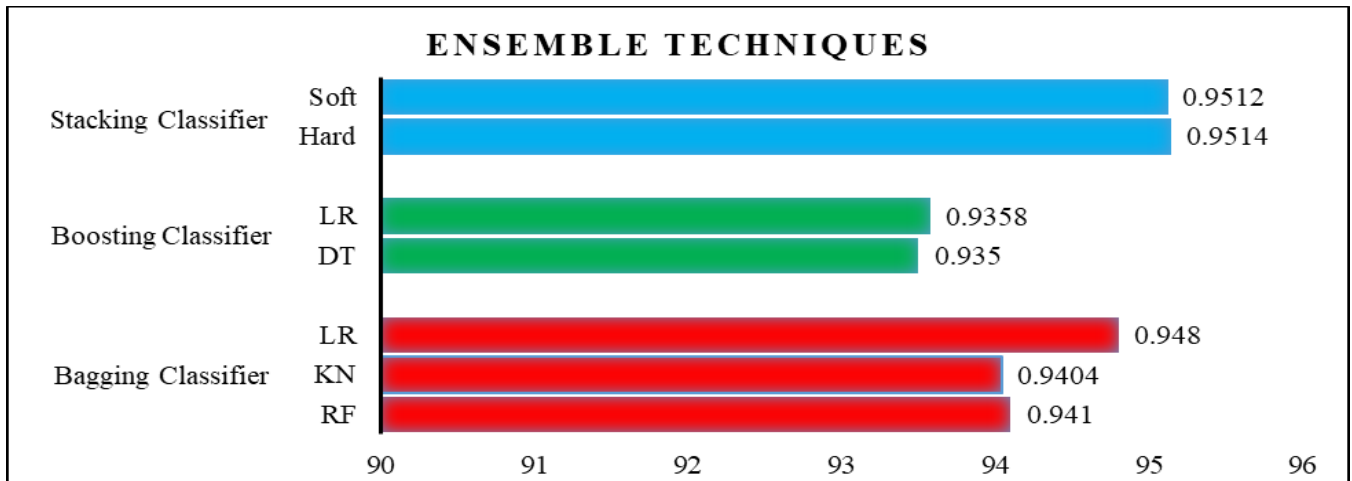
Fig 4 Graphical Representation of Proposed Approach

## 5. Discussion

Users are now more likely to express themselves through the use of multimodal content, such photographs and text, because of the growing popularity of social media and digital platforms. As of now, most hate speech datasets are available in textual form. Considering this aspect, our research focuses on the primarily used textual datasets. We can successfully collect people's opinions regarding particular instances of hate content in our research by using machine learning techniques to interpret the psychological orientation inherent in the information. Our objective differs from the conventional hate speech detection methods. We have contributed to the area of hate speech detection by proposing a novel approach by considering the effect of various preprocessing sequences. Based on our findings, it is evident that sequencing of pre-processing steps matters a lot due to which our method is better able to learn crucial content from twitter datasets and performs robustly.

## 6. Conclusion and Future Directions

Hate speech detection on social media is the most prevalent and ongoing research now a day. To detect the hate content on twitter, this paper implements the ensemble techniques on the best pre-processing sequence. However, the accuracy scores are evaluated for various individual base/weak models on various sets of pre-processing sequences to select the best sequence. This paper also identifies that ensemble techniques performed better than individual classifiers on the proposed sequence to show the effect of preprocess sequencing on evaluation. Out of the various ensemble methods like bagging, boosting and stacking, the stacking (meta-classifier) is identified to be the best and give remarkable results. It can also be concluded that both soft and hard voting classifiers give competition to identify hate speech. As future directions, this method can be implemented on unimodal and multimodal data by considering the sarcasm and sentiments in hate content. Various metaheuristic approaches like ALO and FMO and parametric optimization strategies can be used to tweak various classifier performances.

## 7. Declarations

- *Fundings*

There was no specific grant from a government agency, a business, or a nonprofit group for this research.

- *Conflicts of Interest*

The authors declare that they have no conflict of interest.

- *Consent to participate*

This article does not contain any studies with human participants or animals performed by any of the authors.

- *Consent for Publication*

Not Applicable

- *Code Availability*

Code will be made available on request.

- *Availability for Data and Material*

The datasets used for analyzing are available via the following repository web links:

**WZ-LS -** https://github.com/zeeraktalat/hatespeech
**DT -** https://github.com/t-davidson/hate-speech-and-offensive-language
**FOUNTA -** https://github.com/ENCASEH2020/hatespeech-twitter